\documentclass[runningheads]{llncs}
\usepackage{graphicx}
\usepackage{amsmath,amssymb} 
\usepackage{color}

\begin{document}
\title{Feature2Mass: Visual Feature Processing in Latent Space for Realistic Labeled \newline Mass Generation} 

\titlerunning{Feature2Mass}
%
\author{Jae-Hyeok Lee \and Seong Tae Kim \and Hakmin Lee \and Yong Man Ro{*}}

%
\authorrunning{J. Lee, S.T. Kim, H.M. Lee, and Y.M. Ro}
%


\institute{School of Electrical Engineering, KAIST, Daejeon, Republic of Korea \\
	\email{\{heuyklee, stkim4978, zpqlam12, ymro\}@kaist.ac.kr}}

%
\maketitle              
%
\begin{abstract}
	This paper deals with a method for generating realistic labeled masses. Recently, there have been many attempts to apply deep learning to various bio-image computing fields including computer-aided detection and diagnosis. In order to learn deep network model to be well-behaved in bio-image computing fields, a lot of labeled data is required. However, in many bioimaging fields, the large-size of labeled dataset is scarcely available. Although a few researches have been dedicated to solving this problem through generative model, there are some problems as follows: 1) The generated bio-image does not seem realistic; 2) the variation of generated bio-image is limited; and 3) additional label annotation task is needed. In this study, we propose a realistic labeled bio-image generation method through visual feature processing in latent space. Experimental results have shown that mass images generated by the proposed method were realistic and had wide expression range of targeted mass characteristics.
	
\end{abstract}

\section{Introduction}
Generating realistic labeled bio-images is a highly important task. Recently, there have been many attempts to apply deep learning to computer-aided detection or diagnosis in various bioimaging fields \cite{ref2,ref8,ref12,ref13,ref14}. In order to learn high-performance deep network models, there is immense demand for large amounts of labeled data. However, in many bioimaging fields, the large-size of labeled dataset is scarcely available. Therefore, it is becoming increasingly important to generate realistic labeled data using small amounts of data in bioimaging fields where suffering from lack of labeled data.

Although there have been a few researches dedicated to solving this problem through generative model, there were some limitations as follows: 1) The generated bio-image does not seem realistic \cite{ref3,ref10}; 2) the variation of generated bio-image is limited \cite{ref1,ref5,ref7,ref11}; and 3) additional label annotation task is needed which requires expensive cost \cite{ref6,ref7}. 

In order to overcome aforementioned limitations, we propose a novel realistic labeled bio-image generation method through visual feature processing in latent space. The proposed method learns the generative model with adversarial learning to form manifold in latent feature space using few existing annotated images. After learning the generative model, the encoder of the generative model could map mass images onto the manifold of the latent feature space and we define it as visual feature. The processed visual features, the results of the proposed visual feature processing, could be decoded into pixel space through decoder of the generative model to generate wide expression range of realistic mass image which has targeted characteristics.

The main contribution of this paper is summarized as follows: 1) We have proposed a novel method for generating realistic labeled masses that is not confined to the expression range of the limited real-world data.

\noindent 2) Through the proposed method, we have succeeded in forming an appropriate manifold for the characteristics of masses in latent space which is difficult due to the non-rigid nature of the masses.

\noindent 3) Comprehensive experiments have been conducted to validate the effectiveness of the proposed method. Experimental results show that masses generated by the proposed method are remarkably realistic. Moreover, the generated masses have a wide expression range of targeted mass characteristics.

\section{Generating Realistic Labeled Masses by Visual Feature Processing in Latent Space}
\begin{figure}[b]
	\centering
	\includegraphics[width=10.5cm]{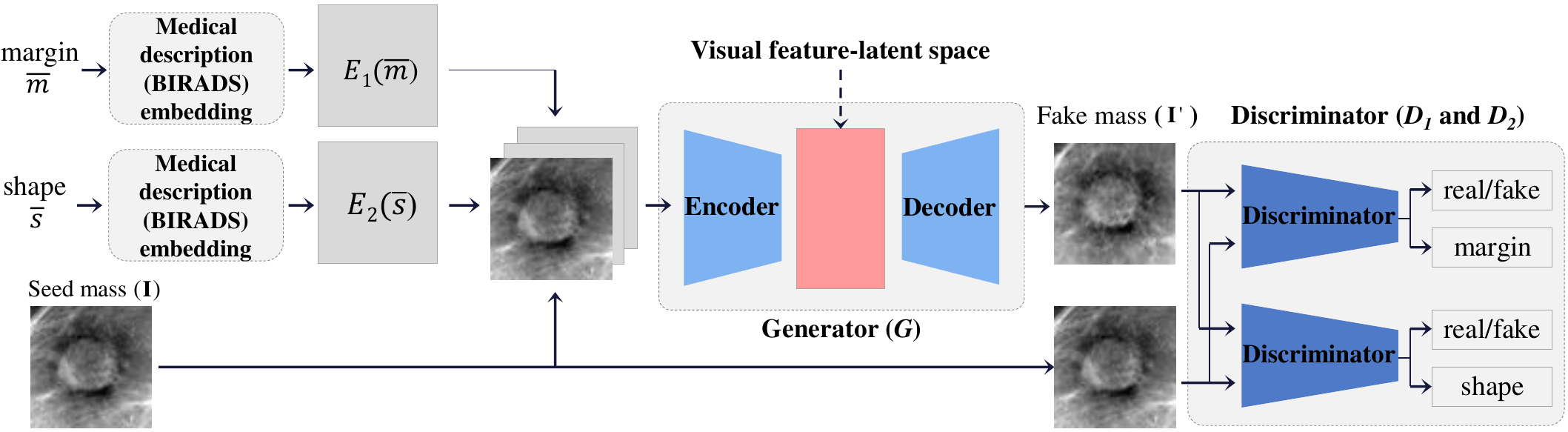}
	\caption{Overall architecture of the proposed method for synthesizing masses through visual feature processing.}
\end{figure}
\subsection{Overview of the Proposed Mass Generation}

In this paper, we design the proposed method to generate breast masses according to the medical description. The BIRADS (Breast Imaging Reporting and Data System) \cite{ref4} which is designed to characterize the masses on breast imaging is used as the medical description in this paper.

Overall architecture of the proposed method is shown in Fig. 1. As seen in Fig. 1, the proposed architecture is built upon Generative Adversarial Networks and it includes four main modules: a mass generator (autoencoder) module ($G$), two discriminator modules ($D_1$ and $D_2$), and BIRADS description embedding modules ($E_1$ and $E_2$). The two discriminators scheme is adopted to effectively extract and backpropagate the BIRADS description characteristics of each generated mass.

The goal of learning phase is to let the $G$ learns the manifold of the breast masses which will be used in generating phase. Overall explanation of the learning/generating phase procedure of the proposed architecture is as follows.

In the learning phase, as seen in Fig. 1, breast mass image and two of the major information in BIRADS, margin ($\bar{m}$) and shape ($\bar{s}$) labels are inputted to the network. The margin and shape labels are embedded into $E_1(\bar{m})$ and $E_2(\bar{s})$ through BIRADS description embedding modules before inputted to $G$. Then $G$ generates the fake mass ($I^\prime$) using the inputted mass ($I$) and embedded labels as
\begin{equation}
	I^\prime = G(I, E_1(\bar{m}), E_2(\bar{s}))),
\end{equation}
\noindent where $E_1(\bar{m})$ and $E_2(\bar{s})$ denotes the embedded margin ($\bar{m}$) and shape ($\bar{s}$) labels of BIRADS description.

In the generating phase, as seen in Fig. 2, we use the $G$ and BIRADS description embedding modules which are trained to approximate breast mass manifold. $G$ includes encoder and decoder which consist of convolution layer and transposed convolution layer, respectively. In front of $G$, a seed breast mass and corresponding margin or shape labels are embedded through BIRADS embedding modules similar to learning phase. When the breast mass with embedded labels come into the input of $G$, the encoder maps it on the latent space formed by the seed mass and corresponding BIRADS description. By performing the feature processing on the latent space, realistic breast masses could be generated by decoding processed visual features into pixel space.
\begin{figure}[t]
	\centering
	\includegraphics[width=10.5cm]{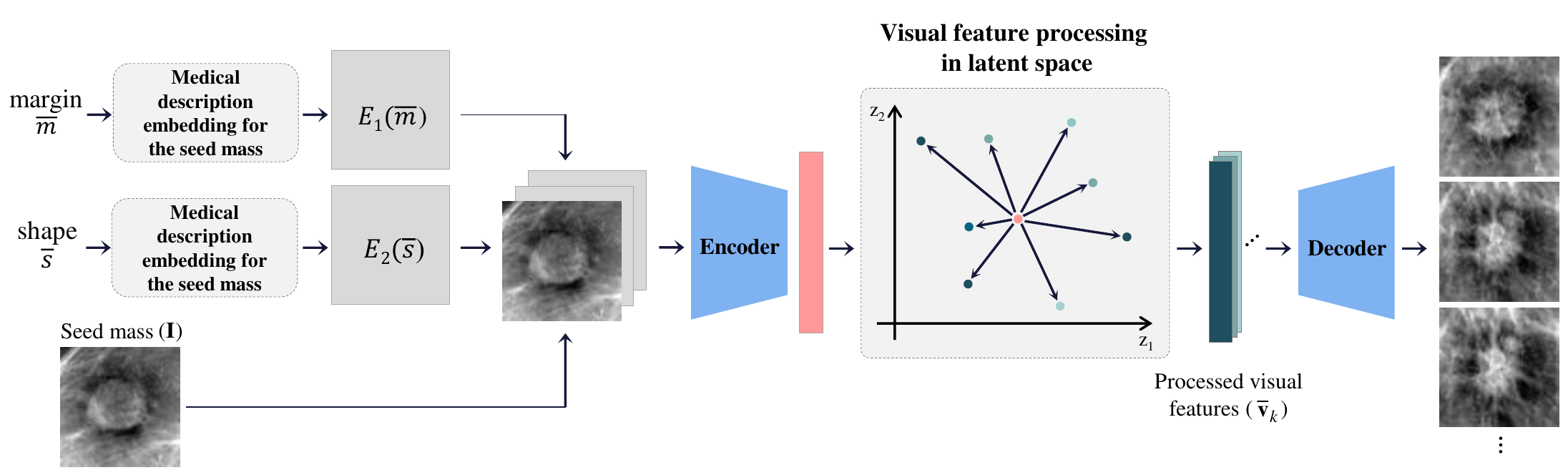}
	\caption{Visual feature processing in the latent space for mass generation using the proposed method.}
\end{figure}
\subsection{Visual Feature Processing}
As described before, the visual feature ($\bar{v}$) can be encoded by mapping the breast mass and embedded labels (medical description) into the latent space through encoder. If a visual feature in the latent space is fed to the decoder, a breast mass representing the visual feature is generated. The generated breast mass contains characteristics of seed breast mass and embedded medical description. To increase the non-linear diversity in mass generation, we devise visual feature processing to consider multitudinous possible visual features on the manifold where visual features of the breast masses exist. Note embedded labels (medical description) and seed mass fix the visual features within the latent space. 

\subsubsection{Visual Feature Processing by Interpolation.}
The various masses are generated from multitudinous possible visual features on the manifold. A visual feature can be obtained by interpolation processing with adjacent fixed visual features. Namely, a new visual feature is acquired through interpolated between adjacent visual features which are fixed by embedded labels (medical description) and seed mass. Interpolation between $N$ different visual features allows considering possible visual features within the range of N visual features.
\begin{equation}
	\begin{gathered}
		\bar{v}_{\text{interpolated}} = \sum_{k=1}^N \alpha_k\bar{v}_k, \\
		\text{where}\ 0\le \alpha_k \le 1 \ \text{and}\ \sum_{k=1}^{N} \alpha_k = 1.
	\end{gathered}
\end{equation}
\noindent $\alpha_k$ denotes the weight multiplied by the visual feature $\bar{v}_k$ in the interpolation. If these $N$ visual features are linearly independent, that is, if the only condition for making a linear combination of visual features 0 is that all $\alpha_k$ is zero, then these $N$ visual features form ($N-1$)-dimension hyperplane. In other words, when $N$ is less than the dimension of the visual feature itself, if $N$ visual features are linearly independent of each other, then all possible visual features in the hyperplane of ($N-1$)-dimension can be considered.

\subsection{Mass Generation through Visual Feature Processing}
In this section, we describe in detail deep learning procedure how the breast mass is generated through the visual feature processing. In the learning procedure, the $G$ learns the manifold of the breast masses which could be used in generating phase. In the generating phase, $G$ and BIRADS description embedding modules are used to generate realistic breast masses.

The behavior of BIRADS description embedding modules in learning and generating phase is as follows. In the learning and generating phases, the BIRADS description embedding modules map one-hot label form of BIRADS description label into the size of the breast mass image. Then the breast mass image and the embedded labels are concatenated and then inputted into the encoder.

The details of each generation step in the generating phase are as follows: 1) The encoder receives the concatenated input and maps it into the 1024-dimensional latent feature space. 2) The aforementioned processing to the visual feature is applied to generate a visual feature of the breast mass with appearance and characteristic that is not presented in the seed breast mass image. In feature processing in learned latent feature space, there is room for expansion since visual feature processing is able to include any operations applicable to visual features besides interpolation. 3) When a processed visual feature is inputted to the decoder, the decoder maps it into the pixel.

\subsection{Learning Strategy of the Proposed Deep Network Framework}
In this study, the proposed deep network framework utilizes two BIRADS description labels (\textit{i.e.} margin and shape), and it has two discriminators ($D_1$ and $D_2$) to predict them. The loss function of $D_1$ is defined as
\begin{equation}
	\begin{gathered}
		L_{D_1} = E_{I \sim P_{\text{data}}}[\log(D_1^{RF}(I)) + \log(1-D_1^{RF}(I^\prime))\\
		+ \lambda_1(\log(1-(\bar{m}-D_1^m(I))) + \log(1-(\bar{m}-D_1^m(I^\prime))))],
	\end{gathered}
\end{equation}
\noindent where the $D_1^{RF}(\cdot)$ and $D_1^{m}(\cdot)$ denotes the prediction about real/fake and estimated margin label of $D_1$, respectively. $\lambda_1$ is the weight multiplied by each loss term to balance overall loss function. The first two terms of loss function represent the general GAN loss of adversarial learning that predicts the real/fake of the real breast mass image ($I$) and generated breast mass image ($I^\prime$). The loss terms in $\lambda_1(\cdot)$ decrease when $D_1$ predicts the ground truth margin label ($\bar{m}$) more precisely from the inputted $I$ and $I^\prime$.

The loss function of $D_2$ is defined as
\begin{equation}
	\begin{gathered}
		L_{D_2} = E_{I \sim P_{\text{data}}}[\log(D_2^{RF}(I)) + \log(1-D_2^{RF}(I^\prime))\\
		+ \lambda_2(\log(1-(\bar{s}-D_2^s(I))) + \log(1-(\bar{s}-D_2^s(I^\prime))))],
	\end{gathered}
\end{equation}
\noindent where $D_2^{s}(\cdot)$ denotes the prediction about shape label of $D_2$. The first two terms of loss function represent the general GAN loss of adversarial learning that predicts the real/fake of the $I$ and $I^\prime$ by $G$. The loss terms in $\lambda_2(\cdot)$ decrease when $D_2$ predict the ground truth shape label ($\bar{s}$) from the inputted $I$ and $I^\prime$.

The loss term that predicts the BIRADS description label by taking $I^\prime$ as an input serves as noise in the early learning phase when $G$ does not generate a realistic breast mass. However, after $G$ has been able to generate a breast masses that is similar to real breast masses, it pushes $D$ to predict the BIRADS description label better for the generated masses which has non-linearly different aspect (but does not differ much enough to have different margins and shape labels). Therefore, it makes $D_1$ and $D_2$ have a data augmentation effect. 

Next, the loss function of $G$ is defined as
\begin{equation}
	\begin{gathered}
		L_{G} = E_{I \sim P_{\text{data}}}[\log(D_1^{RF}(I^\prime)) + \log(D_2^{RF}(I^\prime))\\
		+ \lambda_3(\log(1-(\bar{m}-D_1^m(I^\prime))) + \log(1-(\bar{s}-D_2^s(I^\prime)))) + \lambda_4 \cdot L1(I, I^\prime)],
	\end{gathered}
\end{equation}
\noindent where $\lambda_3$ and $\lambda_4$ are the weights multiplied by each loss term to balance overall loss function. The loss term $L1(I, I^\prime)$ denotes reconstruction loss using L1-norm between real breast mass $I$ and generated breast mass $I^\prime$. The first two loss terms in loss function, like the loss functions of $D_1$ and $D_2$, mean the general GAN losses of adversarial learning for real/fake predictions of $D_1$ and $D_2$ for the $I^\prime$. The loss terms in $\lambda_3(\cdot)$ decrease when $G$ could generate $I^\prime$ that has more strong characteristics along BIRADS description  $\bar{m}$, $\bar{s}$. Like the loss term in $D$, after $D$ has been able to properly identify the BIRADS description label, it pushes $G$ represents the characteristic of the BIRADS description label better which is used in $I^\prime$ generation. The last $L1$ loss term pushes $G$ to form a manifold similar to the real data distribution.

\section{Experimental Results}
\subsection{Dataset}
For identifying the effectiveness of the proposed method, we utilized the publicly available DDSM dataset. The mammograms scanned by Howtek 960 were selected from DDSM dataset for the experiments. A total of 841 regions of interest (ROI) were used. The size of seed image $I$ was resized to 64 by 64. For the BIRADS description, as aforementioned, shape and margin of masses were selected from BIRADS descriptions. Since these are representative characteristics of breast masses and widely used for recording in clinical reports.

\subsection{Architecture and Training Details}
Each BIRADS description embedding module consisted of two fully-connected layers which have 256 and 4096 neurons. The encoder and decoder of the proposed architecture composed of seven convolution and transposed convolution layers, respectively. The discriminator module composed of ten convolution layers and three fully-connected layers. Each module utilizes LeakyReLU-Conv (or Transposed Conv)-BatchNorm structure.

For training the generator and discriminator, the Adam optimization was used with learning rate 0.0002 and pytorch default Adam optimizer settings. The values of loss function balancing weights $\lambda_1$, $\lambda_2$, $\lambda_3$, and $\lambda_4$ were 10, 10, 10, and 300, respectively. We utilized a batch size of 512 and trained the network for 8000 epochs. For image data augmentation, horizontal flipping, vertical flipping, and cropping were performed in a random manner.

\subsection{Visual Feature Processing in Latent Space with Interpolation}
\subsubsection{Visual Feature Interpolation between Two Visual Features.}
The visual feature interpolation results between two visual features are shown in Fig. 3. The two visual features were selected from 841 visual features of 841 mass ROIs in the dataset. In experiments, two visual features were selected and equidistant visual features between the two features were interpolated. The interpolated visual features were decoded into pixel space. The leftmost and rightmost images in Fig. 3 are two seed breast mass images. Among the images in the middle, the leftmost and rightmost images represent decoded images from visual features fixed with seed image and embedded labels. The eight images that exist between them are generated images through the proposed visual features interpolation. 


\subsubsection{Visual Feature Interpolation from Three Visual Features.}
The visual feature interpolation results from three visual features are shown in Fig. 4. The three visual features were selected from 841 visual features of 841 mass ROIs in the dataset. As seen in Fig. 4, twelve interpolated visual features from the three visual features were visualized. By inputting twelve visual features into the decoder and mapping them to pixel space, we verified that the manifold from these three seed breast masses was formed suitably.
\begin{figure}[t]
	\centering
	\includegraphics[width=11cm]{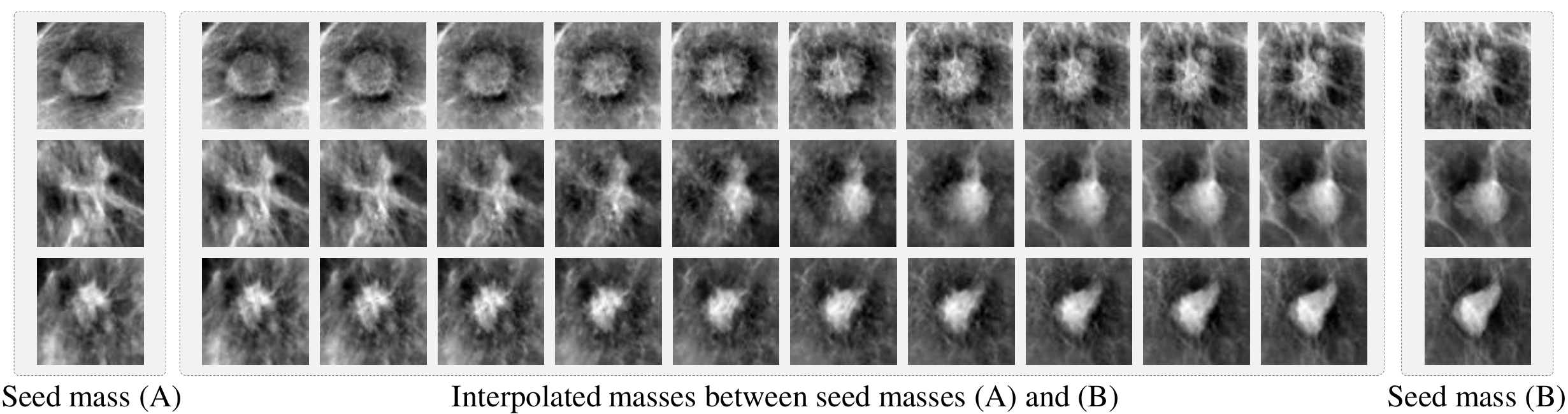}
	\caption{Generated breast masses through the proposed visual feature interpolation between two visual features.}
\end{figure}
\begin{figure}[t]
	\centering
	\includegraphics[width=6.3cm]{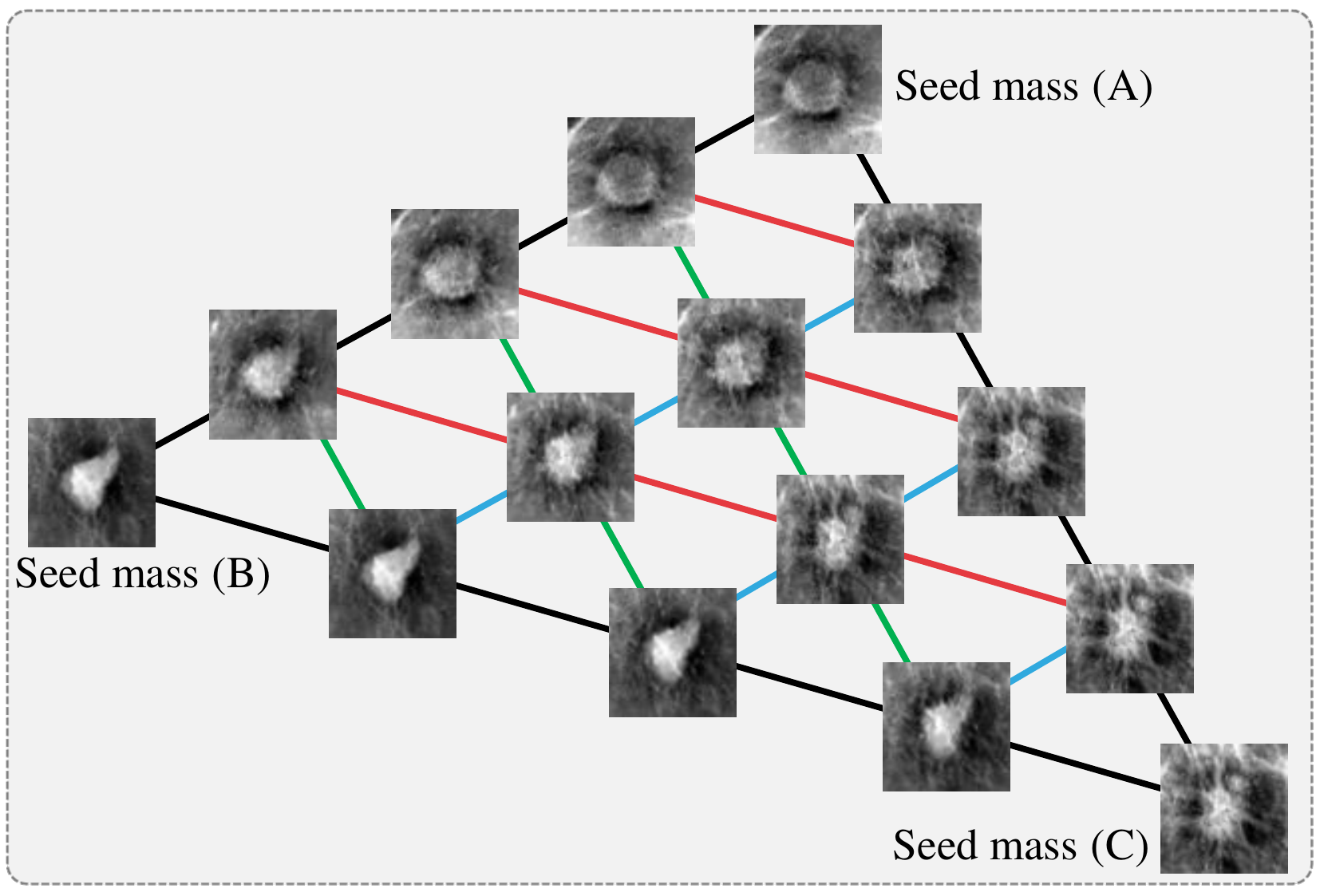}
	\caption{Generated breast masses through the proposed visual feature interpolation from three visual features.}
\end{figure}
\subsection{Visual Feature Processing within a Specific BIRADS Category}
This section demonstrates that generating masses with intended annotation information is achievable through the proposed method. Fig. 5 shows the generated masses using visual features obtained from the seed masses with a specific BIRADS category ($C$) (\textit{e.g.} ill-defined margins and round shape). The total number of seed masses used in the experiment was 161. Among them, 20 masses had ill-defined margins and round shape, 104 masses had spiculate margins and irregular shape, and 37 masses had circumscribed margins and oval shape.

As seen in Fig. 5, the masses in the left side of corresponding three seed masses were generated from interpolated visual features. The interpolated visual feature was calculated as follows: 1) Twenty visual features in a specific BIRADS category ($\bar{v}_\text{sel}^C$) were randomly selected out of the number of candidate masses in a specific category; 2) The corresponding twenty weights ($\alpha_\text{sel}^C$) were randomly initialized in the unit of 0.05.

In Fig. 5, the top three masses which have the largest weights $\alpha_\text{sel}^C$ and corresponding weights are represented in the right side of each generated mass. As seen in Fig. 5, the generated masses were realistic and had target characteristics (\textit{e.g.} ill-defined margins and round shape). Therefore, the masses which are generated utilizing visual features in a specific BIRADS description category did not require additional labeling cost.
\begin{figure}[t]
	\centering
	\includegraphics[width=11cm]{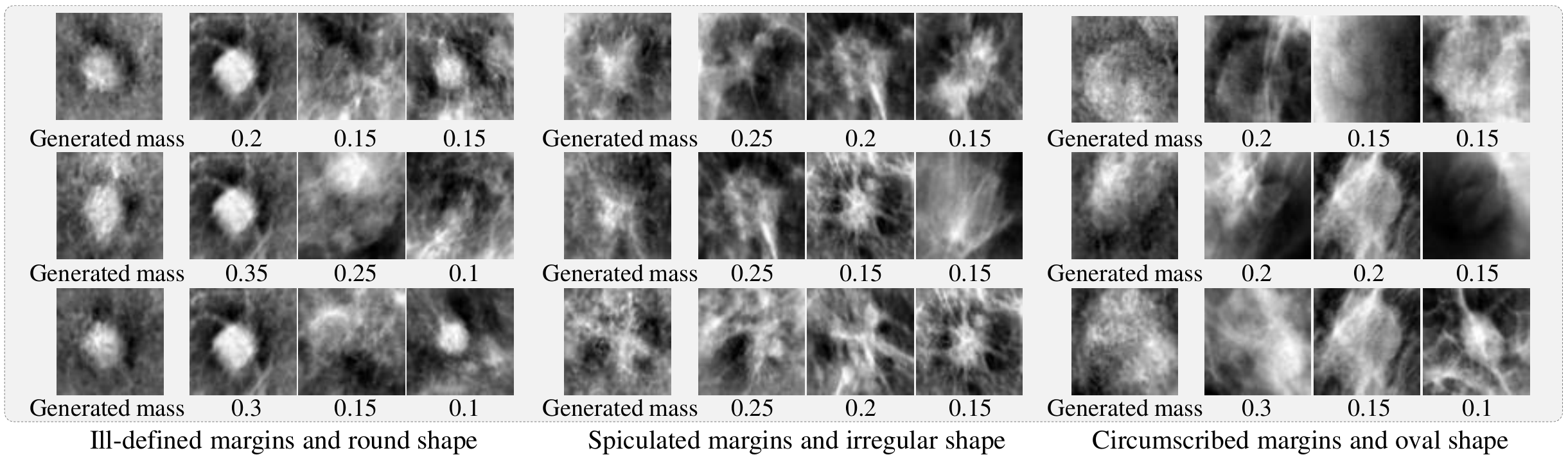}
	\caption{Generated breast masses through the proposed visual feature processing with visual features belonging to a specific BIRADS category. The seed breast masses have ill-defined margins and round shape (left), speculated margins and irregular shape (middle), and circumscribed margins and oval shape (right).}
\end{figure}
\section{Conclusions}
In this paper, we proposed the novel bio-image generation method through visual feature processing. The proposed method learned the generative model with adversarial learning to effectively form manifold in latent feature space using a limited number of annotated mass images. After learning the generative model, the encoder of the generative model could map mass images onto the manifold of the latent feature space (defined as visual feature). By decoding the processed visual features, the mass image was generated. Through extensive experiments, we verified that the masses generated by the proposed method were realistic and had a wide expression range. Moreover, it was possible to generate masses with the target characteristics. By generating the masses with the target characteristics, it could alleviate the labeling workload for utilizing generated masses in real-world. It is expected that the proposed method could be generalized to other bioimaging fields where suffering from lack of annotated data.

\subsubsection{Acknowledgement.} This work was supported by Institute for Information \& communications Technology Promotion (IITP) grant funded by the Korea government (MSIT) (No. 2017-0-01778, Development of Explainable Human-level Deep Machine Learning Inference Framework). {*}Y.M. Ro is a corresponding author.

\clearpage
\bibliographystyle{splncs04}
\bibliography{egbib}
\end{document}